\documentclass[letterpaper, 10 pt, conference]{ieeeconf}  

\usepackage{times}
\usepackage[numbers]{natbib}
\usepackage{multicol}
\usepackage[bookmarks=true]{hyperref}

\usepackage{lipsum}
\usepackage{comment}
\usepackage{balance}
\usepackage{graphicx}
\usepackage[table]{xcolor}
\usepackage{caption}
\usepackage{listings}
\usepackage{xcolor}

\definecolor{codegreen}{rgb}{0,0.6,0}
\definecolor{codegray}{rgb}{0.5,0.5,0.5}
\definecolor{codepurple}{rgb}{0.58,0,0.82}
\definecolor{backcolour}{rgb}{0.95,0.95,0.92}

\lstdefinestyle{mystyle}{
    backgroundcolor=\color{backcolour},
    commentstyle=\color{codegreen},
    keywordstyle=\color{magenta},
    numberstyle=\tiny\color{codegray},
    stringstyle=\color{codepurple},
    basicstyle=\ttfamily,
    breaklines=true,
    captionpos=b,
    keepspaces=true,
    numbers=left,
    numbersep=5pt,
    showstringspaces=false,
    tabsize=2
}
\usepackage{minted}
\usepackage{xcolor}
\definecolor{bg}{HTML}{f2f2ea}
\usepackage{float} 

\usepackage{seqsplit}

\IEEEoverridecommandlockouts                              

\usepackage{tabularx}
\usepackage{multirow}

\usepackage{amssymb}  
\usepackage[font=small]{caption}

\title{\Large \bf
An Analysis of Streaming Deep Reinforcement Learning for \\ Adaptive Continual Learning in Robotics}

\author{Teeratham Vitchutripop$^{1}$, Alyssa Quarles$^{1}$, Wenhe Zhang$^{1}$, Richard Xue$^{1}$, Daniel Rakita$^{1}$
\thanks{$^{1}$Department of Computer Science, Yale University, New Haven, CT 06520, USA}}

\begin{document}

\maketitle
\thispagestyle{empty}
\pagestyle{empty}


\begin{abstract}

Over the course of a lifetime, robots may encounter novel scenarios unaccounted for in its original training that result in performance degradation. One common approach to mitigating this issue is to further grow the offline training dataset in hopes of producing a policy robust to these changes. In contrast, biological learning occurs moment-to-moment via a \textit{stream of experience}, unlike the predominantly batch-based and offline nature of deep learning. Although recent works show the feasibility of \textit{stream-based deep reinforcement learning}, where updates use only the latest experience, none have shown it to be a viable continual learning framework for adapting robotic policies to unseen changes. In this paper, we present the first analysis of streaming deep reinforcement learning for adaptive continual learning in robotics. In particular, we show that, following an initial pretraining phase, streaming deep RL can enable a robot to successfully adapt to unforeseen changes to itself, its environment, or goals. Our primary experiments within quadruped locomotion demonstrate that a deep neural network robotic policy with certain optimizers and plasticity loss mitigation techniques can successfully leverage domain task knowledge from its pretraining to quickly adapt online to diverse changes via stream learning, outperforming batch-based on-policy methods and improving task success rates by up to 90\% over the pretrained policy. Furthermore, we perform additional evaluations on robotic manipulation tasks to determine if our previous observations extend to different robotic morphologies and scenarios. Our results show that the successes observed in quadruped locomotion can be partially realized in manipulation with stability and performance limitations. We conclude with a discussion on the limitations of our work and its implications for the future of continual robot learning.
\end{abstract}








\section{Introduction}
\label{sec:introduction}
The capacity for robots to continuously learn from their own stream of experience is attractive both for its parallels to biological learning and the fast adaptation it could enable. Even as computational resources expand, the ``world'' in a learning problem remains orders of magnitude larger and more complex than the agent, making it impractical to assume pretrained models can anticipate every scenario \cite{javed2024the}. This observation motivates an alternative paradigm: designing agents that may be smaller and more resource constrained, but can continually learn online via a \textit{stream} of experience (i.e., stream learning) to meet the decision-making problems of the present. This strategy mirrors how humans and animals, while limited by the size and energy of a biological brain, still possess a remarkable capacity to learn moment-to-moment through experience \cite{Hayes2021-ti}.


Robots through stream learning could adapt to scenarios that would otherwise be catastrophic for a fixed policy. For instance, a quadruped with a damaged leg, trained under the assumption of full functionality, would normally be incapacitated until repaired. However, in an isolated environment far from assistance, a sufficiently plastic policy capable of learning from a stream of experience could enable the robot to adapt and continue walking.


\begin{figure}
    \centering
    \includegraphics[width=0.95\linewidth]{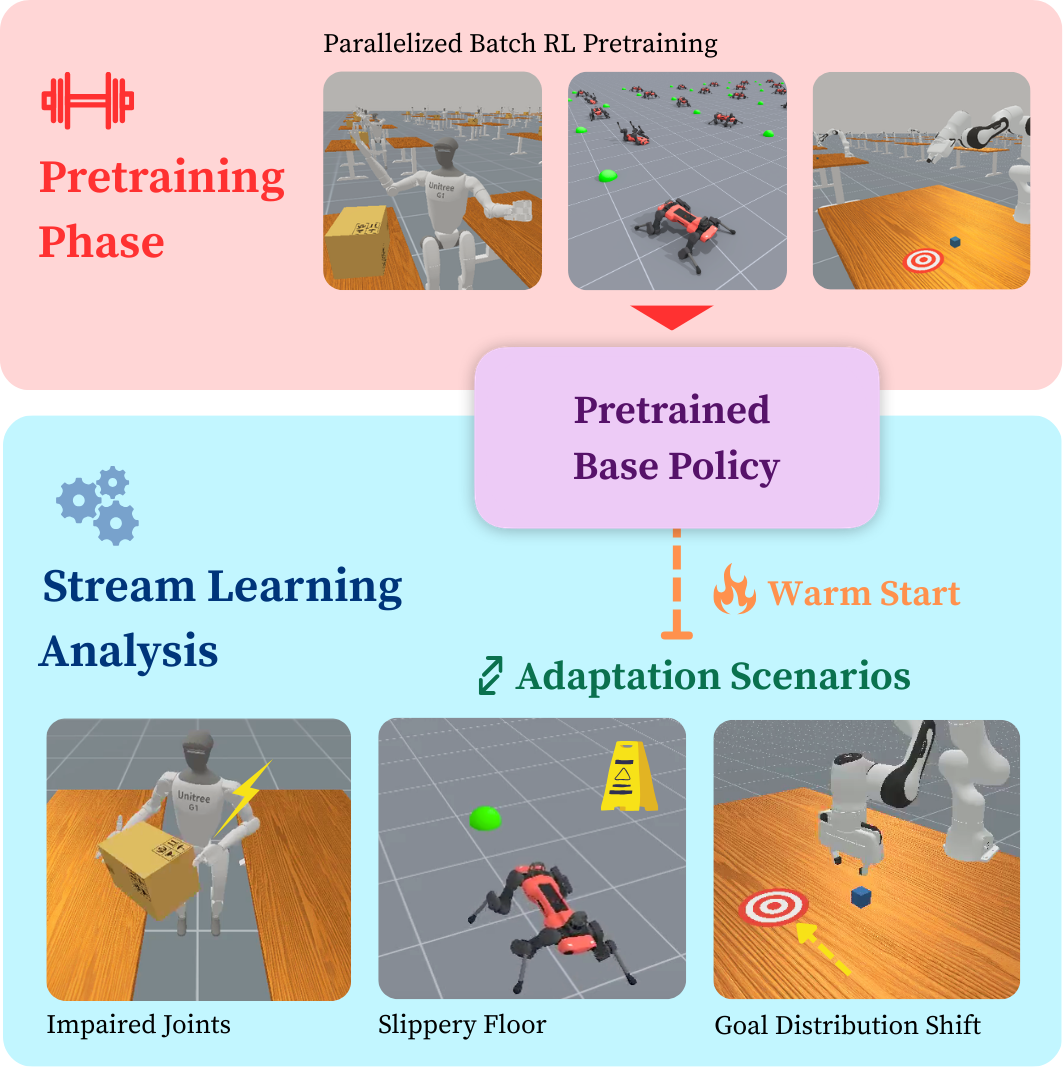}
    \caption{An overview of our analysis on streaming deep reinforcement learning for continual adaptation in robotic systems. We examine if pretrained RL policies are able to adapt online in a stream-fashion with no replay buffers to a set of scenarios that introduce a sudden change to the robot, its environment, or goals.}
    \label{fig:overview}
    \vspace{-2em}
\end{figure}

Despite its promise, stream learning contrasts with the dominant deep learning literature in robotics. State-of-the-art approaches, e.g., vision-language action models \cite{intelligence2025pi05visionlanguageactionmodelopenworld}, large behavior models \cite{octo_2023}, and other imitation learning approaches, generally assume batch-based offline learning focused on scaling model size and datasets. Likewise, deep reinforcement learning (RL) has moved away from classical stream-based RL~\cite{10.5555/3312046}, favoring batch updates and replay buffers to improve stability and sample efficiency~\cite{andrychowicz_hindsight_2017}.


Recently, however, stream learning has begun to re-emerge within deep RL~\cite{elsayed2024streamingdeepreinforcementlearning, vasan_deep_2024}. This progress has been primarily spurred by advances in optimizers and by scaling and normalization techniques inspired by neural network plasticity research, which examines how networks maintain performance while learning from new data over time \cite{pmlr-v202-lyle23b}. Despite this exciting progress, no prior work has established whether stream learning using these advancements can serve as a viable continual learning framework for adapting robotic policies to unseen changes.



In this paper, we present the first analysis of streaming deep reinforcement learning for adaptive continual learning in robotics (Fig.~\ref{fig:overview}). Our analysis particularly focuses on adapting pretrained policies, leveraging the advancements made in parallelized batch RL to establish a base policy. We begin by studying how streaming deep RL can enable successful adaptation to unforeseen changes to the robot, its environment, or goals within a quadruped locomotion domain. More specifically, we examine the role that optimizers and plasticity loss mitigation techniques play in enabling a robot to recover from abrupt changes and maintain robust performance over time. We follow up by conducting additional evaluations in object manipulation tasks to determine if the observations made within quadruped locomotion generalize to other robotic morphologies and scenarios.


Our results suggest that pretrained deep RL policies can maintain performance after switching to stream learning and successfully adapt to novel scenarios. We observe that deep neural networks remain sufficiently malleable and stable during online stream learning, allowing a robot to autonomously self-improve from its experiences by up to 90\% in success rate over the base policy on certain adaptation scenarios. Notably, the top four performing streaming deep RL variants outperform batch-based on-policy methods by 60\% in max success rate on average across our adaptation scenarios. We identify that optimizer choice contributes significantly to stability and that plasticity loss mitigation techniques like layer normalization \cite{ba2016layernormalization} and continual backpropagation \cite{dohare_loss_2024} can be helpful in enhancing adaptation performance and recovery speed. Furthermore, we find that streaming deep RL can be effective on manipulation tasks with proper tuning, but suffers from performance decline over time and limited recovery in more complex settings. We conclude with a discussion on the limitations of our work and potential opportunities for future research stemming from our analyses.

We release our code as open source to enable the robotics community to evaluate, reproduce, and build upon our work.\footnote{\href{https://github.com/tjvitchutripop/stream-rl-robotics/}{Link to GitHub repository.}}

\section{Related Works}
\label{sec:related_works}
\subsection{Batch and Stream Reinforcement Learning}

Foundational RL algorithms such as TD, SARSA, and Actor-Critic \cite{10.5555/3312046} were designed for stream learning, updating policies incrementally from each experience without storage. Their scope, however, was largely limited to tabular settings. With the rise of deep neural networks as powerful function approximators, these methods achieved breakthroughs in video games \cite{mnih_human-level_2015} and continuous control \cite{andrychowicz_hindsight_2017}. Yet, as deep RL matured, batch-based practices from deep learning took hold: replay buffers \cite{andrychowicz_hindsight_2017}, offline RL \cite{levine2020offlinereinforcementlearningtutorial}, and GPU-enabled parallel simulation \cite{taomaniskill3} all shifted the field away from streaming. This shift was driven both by the sample efficiency of reusing costly experience and by the training instabilities of stream learning with neural networks, which \citet{elsayed2024streamingdeepreinforcementlearning} terms the \textit{stream barrier}.


To address the stream barrier, \citet{Elfwing2017SigmoidWeightedLU} proposed a sigmoid-weighted activation function with eligibility traces. More recent work shows that normalization and scaling techniques, such as observation and layer normalization, can greatly improve streaming deep RL \cite{elsayed2024streamingdeepreinforcementlearning, vasan_deep_2024}. \citet{vasan_deep_2024} introduced a normalization-based policy gradient method with RG estimation, while \citet{elsayed2024streamingdeepreinforcementlearning} identified large step sizes as a key instability source and proposed Overshooting-bounded Gradient Descent (ObGD), an optimizer using eligibility traces.



While \citet{vasan_deep_2024} successfully applied streaming deep RL to train robotic systems from scratch (e.g., manipulator reaching and Roomba navigation), certain tasks like quadruped locomotion or object manipulation are more challenging and inefficient to learn from scratch via a stream-setting. Thus, our work leverages the advancements made in parallelized simulation and batch RL to bootstrap the policy via pretraining. Moreover, no prior work has explored streaming deep RL's potential for adaptation and continual learning in robotics.



\subsection{Plasticity in Deep Learning}
Plasticity refers to a neural network's ability to continue adapting as it encounters new data \cite{pmlr-v202-lyle23b}. For continual and streaming deep RL, plasticity is crucial: policies must remain flexible under distribution shifts and ongoing updates. However, prior work shows that standard deep neural networks often lose this capacity over time, a phenomenon known as \textit{plasticity loss}~\cite{dohare_loss_2024, nikishin_deep_2023, muppidi2024fast}.  

Several strategies have been proposed to counter plasticity loss, including reinitializing underused units \cite{dohare_loss_2024}, injecting new parameters \cite{nikishin_deep_2023}, adjusting optimizers \cite{muppidi2024fast}, and applying normalization layers \cite{pmlr-v202-lyle23b}. These techniques highlight that preserving plasticity is as much about stabilizing optimization as it is about maintaining network expressivity. Our work revisits this issue in the context of streaming deep RL, analyzing whether current methods allow pretrained policies to adapt reliably to new robotic scenarios, rather than just a single fixed task.

\subsection{Continual Learning and Adaptation in Robotics}
Continual learning is a machine learning paradigm where models adapt to shifts in data distributions and objectives over time \cite{lesort_continual_2020}. In robotics, many approaches enable adaptation of pretrained behavior cloning policies. For example, \citet{ankile2024imitationrefinementresidual} uses an online residual RL policy to correct a frozen imitation model. \citet{sun2025dynamic} introduces dynamic rank adjustment for adapting pre-trained diffusion policies with limited human corrections for online imitation learning.

Beyond behavior cloning, several methods target RL policies directly. \citet{DBLP:conf/rss/LyuLZZDW24} adapt pretrained RL policies with torque compensation to handle dynamic disturbances. \citet{dong2025batch} propose batch online RL, alternating between offline pretraining, real-world data collection, and retraining. In contrast, our work investigates whether streaming deep RL itself can serve as a foundation for continual adaptation in robotics. Although continual learning is also concerned with catastrophic forgetting (backward transfer), we limit the scope of our analysis to forward transfer, using prior knowledge to learn to adapt to unseen conditions.

\section{Background}

In this section, we provide background for concepts that will be relevant for our analyses in \S\ref{sec:analysis} and \S\ref{sec:analysis2}.

\label{sec:technical_overview}
\subsection{Markov Decision Processes}
We formulate the problem of solving a robotic task as an episodic Markov Decision Process (MDP), given by the tuple $(\mathcal{S}, \mathcal{A}, \mathcal{P}, \mathcal{R}, \gamma)$, where $\mathcal{S}$ is the continuous state space, $\mathcal{A}$ is the continuous action space, $\mathcal{P}$ is the transition probabilities, $\mathcal{R}$ is the normalized dense reward function, and $\gamma \in [0,1]$ is the discount factor. At each timestep $t$, the agent interacts with the environment by sampling an action $A_t$ from a parameterized policy $\pi_{\theta}$ based on the current state $S_t$, $A_t \sim \pi_{\theta}(\cdot|S_t)$. The objective of the agent is to find a weight parameterization $\theta$ for $\pi_{\theta}$ that maximizes the expected return, defined as $J(\theta) = \mathbb{E}_{\pi_\theta}\Big[\sum_{t=0}^{T} \gamma^t R(S_t, A_t)\Big]$, where $T$ is the episode horizon. 

\subsection{Online and Streaming Reinforcement Learning}
One approach to solving an MDP is online training with deep reinforcement learning. On-policy RL methods such as Proximal Policy Optimization (PPO) \cite{schulman2017proximalpolicyoptimizationalgorithms} may use an actor-critic architecture, where the actor network represents the policy $\pi_{\theta}$ and the critic network approximates the value function $V^{\pi}(s) = \mathbb{E}_{\pi_\theta}\!\Big[\sum_{t=0}^{T} \gamma^t R(S_t, A_t) \,\big|\, S_0 = s \Big]$ with parameterized function $V_\phi$. During training, an agent in state $S_t$ samples an action $A_t$ from its policy, receives a reward $R_t$, and transitions to $S_{t+1}$. These transitions $(S_t,A_t,R_t,S_{t+1})$ can be used to update the actor-critic network parameters via temporal-difference (TD) learning using the TD error $\delta = (r_t + \gamma V_\phi(s_{t+1})) - V_\phi(s_{t})$ \cite{10.5555/3312046}. In batch-based on-policy RL, the agent collects $b$ transitions before each update, where $b$ is a batch size hyperparameter. This process repeats until the policy converges.

Our analysis is focused on a subset of online RL approaches known as streaming deep RL. Unlike batch-based RL, at each timestep $t$, streaming deep RL algorithms, like Stream-AC \cite{elsayed2024streamingdeepreinforcementlearning}, update the parameters of $\pi_{\theta}$ and $V_\phi$ by computing the TD error $\delta$ based on only the most recent transition, which in essence is setting $b=1$. Theoretically, this setting should enable the agent to adapt faster to changes, as the policy is continuously updated at each step with new experience. However, stream learning is prone to training instabilities, as large parameter updates to $\theta$ or $\phi$ from a noisy TD error can destabilize the agent's policy, resulting in catastrophic performance collapse and policy divergence. At the same time, overly conservative updates will cause the agent to fail to learn the task. These challenges in streaming deep RL highlight the need for special attention to controlling the size of updates and maintaining the networks' plasticity.


\subsection{Area of Investigation}
\label{sec:area_of_investigation}

Through this work, we are generally assessing a robot's ability to adapt to new scenarios. In particular, we consider settings where the transition dynamics $\mathcal{P}$ and reward function $\mathcal{R}$ of the underlying MDP can change at any time. For example, $\mathcal{P}$ may shift due to modifications in agent morphology or environmental conditions, while $\mathcal{R}$ may change due to variations in the goal distribution. 

While standard single-task RL assumes a policy $\pi_\theta$ trained to perform well in a single, stationary MDP $M = (\mathcal{S}, \mathcal{A}, \mathcal{P}, \mathcal{R}, \gamma)$, our setting is defined by a pair of MDPs in sequence $[M_0, M_1] = [(\mathcal{S}, \mathcal{A}, \mathcal{P}_0, \mathcal{R}_0, \gamma), (\mathcal{S}, \mathcal{A}, \mathcal{P}_1, \mathcal{R}_1, \gamma)]$, where $M_0$ is the same conditions as the pretraining phase and $M_1$ is an adaptation scenario. We assume these transitions occur abruptly, without gradual interpolation between MDPs.


Our goal is to investigate to what extent a policy $\pi_\theta$ can adapt \textit{smoothly and efficiently} across these sharp changes. Ideally, the agent leverages the pretrained parameters $\theta_{0}$, optimized for $M_{0}$, to quickly produce a new set of parameters $\theta_1$ that achieves strong performance in the new MDP $M_1$.

\subsection{Optimizers for Stream Learning}
\label{subsec:optimizers}
In our analysis, we will be assessing different optimization schemes for stream learning. Due to its popularity and ubiquity in deep learning, we use the Adam \cite{2015-kingma} optimizer as a baseline. Adam is, however, attributed as being one of the main sources of instability causing the stream barrier and plasticity loss \cite{elsayed2024streamingdeepreinforcementlearning, pmlr-v202-lyle23b}. To address this issue, \citet{elsayed2024streamingdeepreinforcementlearning} introduce two optimizers that de-emphasize large updates by bounding the effective step size: Overshooting-bounded Gradient Descent (ObGD), and Adaptive Overshooting-bounded Gradient Descent (AdaptiveObGD). We provide a high-level overview of these optimizers below, and refer to the reader to the work of \citet{elsayed2024streamingdeepreinforcementlearning} for full details.

ObGD and AdaptiveObGD use eligibility traces, i.e., short-term memory vectors of decayed parameter sensitivities, to assist with multi-step credit assignment. Eligibility trace vectors $z_t$ are particularly useful in a stream learning setting where experience is used for an update and discarded. For ObGD, given $\bar\delta = \max(|\delta|,1)$, eligibility trace $z_t$, step size $\alpha$, and scaling factor $\kappa$, the step size is bounded and updated via the following rule: $\alpha \leftarrow \min(\frac{1}{\kappa\bar\delta||z_t||_1}, \alpha)$. The parameters $\theta$ and $\phi$ are then updated by $w\leftarrow w + \alpha \delta z_w$, where $w$ is the weights. Note that the actor and critic networks have separate optimizers. AdaptiveObGD slightly modifies the ObGD optimizer to be more like Adam by adding a second moment estimate $v$. This estimate is then used for updating the step size, $\alpha \leftarrow \min(\frac{1}{\kappa\bar\delta||\frac{z_t}{\sqrt{v+\epsilon}}||_1}, \alpha)$, and actor-critic weights, $w\leftarrow w + \alpha \frac{\delta z_w}{\sqrt{v+\epsilon}}$.

\subsection{Methods for Combating Plasticity Loss}
Our analysis also assesses the efficacy of plasticity loss mitigation techniques for stream learning.  In particular, we evaluate two appraoches: layer normalization (LN) \cite{ba2016layernormalization} and continual backpropagation (CBP) \cite{dohare_loss_2024}.  We provide a high-level overview for both methods below, but refer the reader to the original work for full details.

Layer normalization (LN) normalizes values across the feature dimension of the intermediate representations for a single sample during a forward pass within a neural network. This normalization step is commonly applied prior to the activation function of a layer and has been shown to be effective for both combating plasticity loss \cite{pmlr-v202-lyle23b} and streaming deep RL \cite{elsayed2024streamingdeepreinforcementlearning}.

Continual backpropagation (CBP) is an algorithm designed to help preserve the plasticity of networks by reactivating the least used units determined via a metric called the contribution utility. A unit's contribution utility $\mathbf{u}_l[i]$ is the sum of the utility of its outgoing connections and is maintained as a running average with with a decay rate, $\eta$. Although the original algorithm scales the utility with the absolute output of the $i$th hidden unit in layer $l$ at time $t$, $|\mathbf{h}_{l,i,t}|$, as we use a Tanh activation function for our networks, we redefine the utility to favor units that that have lower saturation (i.e., higher potential to change). We replace the activation magnitude, $|\mathbf{h}_{l,i,t}|$, with the gradient capacity, $(1-\mathbf{h}_{l,i,t}^2)$, resulting in the following contribution utility: 
\[\mathbf{u}_l[i] = \eta \times \mathbf{u}_l[i] 
+ (1 - \eta) \times (1-\mathbf{h}_{l,i,t}^2)
\times \sum_{k=1}^{n_{l+1}} |\mathbf{w}_{l,i,k,t}|,\]
where $\sum_{k=1}^{n_{l+1}} |\mathbf{w}_{l,i,k,t}|$ is the sum of weights extending from this particular unit to the $n_{l+1}$ units in layer ${l+1}$. 

To prevent the resetted weights from being immediately reinitialized, a maturity threshold hyperparameter $m$ protects them for $m$ steps. The fraction of units replaced at each step is also controlled by a hyperparamter $\rho$, though this value is usually set to be very small, such that one unit is not replaced until several updates occur.

\section{Analysis on Quadruped Locomotion}
\label{sec:analysis}
Our primary analysis is a series of experiments conducted within quadruped locomotion that 1) demonstrate the potential capabilities of streaming deep RL for enabling continual robotic adaptation and 2) analyze how the choice of certain optimizers or plasticity loss mitigation techniques can affect adaptation performance. 

\subsection{Experimental Setup}
\label{sec:analysis-setup}
For all policy training, PyTorch was used as the machine learning framework. The primary metric used for evaluating a policy's performance on a task is success rate, which is measured by performing inference with the current policy across 50 episodes with or without the effects of the adaptation scenario, depending on whether the adaptation phase has begun. Note that no parameter updates occur during these evaluations. We evaluate the policy's success rate every 10,000 steps during online training. All training runs were conducted with five random seeds. 

All comparative methods, except AdaptiveObGD + LN, use the same set of pretrained weights from a successful parallelized batch PPO training run on the base task. AdaptiveObGD + LN uses a set of weights that incorporates layer normalization in the pretraining. A 4-layer multi-layer perceptron with size 256 hidden layers and a Tanh activation function were used for the actor-critic networks. This architecture is consistent across all experiments.

\subsection{Task Environment}
Our analysis is performed within the AnymalC-Reach-v1 task environment in ManiSkill3 \cite{taomaniskill3}, which requires an Anymal C quadrupedal robot to walk to a goal position in the environment. The robot successfully completes the task when it is within 0.35 meters of the goal. The goal is randomized within a certain distribution in front of the robot. A state space comprised of joint position and velocity, linear and angular root velocity, the 2D goal position, and success boolean is used for observations. Actions are represented by the change in joint positions, thus requiring the policy to learn to walk with the quadruped. We use the provided normalized dense reward function, which includes a basic distance-to-goal reward and an assortment of penalties that discourage undesired contacts and high velocities to shape the robot's walking behavior.

Before any adaptation scenarios are introduced, we have a warm-start period where the task environment remains unchanged from the standard version. This stage serves as $M_0$ in a sequence that forms the non-stationary MDP described in \S\ref{sec:area_of_investigation}, mimicking a scenario where a robot is deployed initially in a familiar setting before needing to adapt to any abrupt changes. The warm-start period is 500,000 steps for all experiments which include it.

\subsection{Adaptation Scenarios}

We developed 3 adaptation scenarios that test the robot's ability to adapt to changes to its embodiment, environment, and goals. \textbf{(1) Broken Leg}: We zero-mask 3 values of the action space to simulate the back left leg of the robot being stuck. \textbf{(2) Slippery Floor}: We change the static and dynamic friction coefficients of the ground in the environment from 0.3 to -1.8. \textbf{(3) Goal Shift}: We offset the distribution of goal positions from the original centered at (0,0), to (0,3.0).

\begin{table*}[ht]
\centering
\caption{Table 1. Average Success Rates $\pm$ STD Across 5 Seeds on Quadruped Locomotion Adaptation Scenarios.}
\begin{tabular}{l|cc|cc|cc}
\hline
\multirow{2}{*}{\textbf{Method}} & \multicolumn{2}{c|}{\textbf{Broken Leg}} & \multicolumn{2}{c|}{\textbf{Goal Shift}} & \multicolumn{2}{c}{\textbf{Slippery Floor}} \\
 & Max & Mean & Max & Mean & Max & Mean \\
\hline
Adam & 0.188 $\pm$ 0.027 & 0.080 $\pm$ 0.017 & 0.108 $\pm$ 0.018 & 0.026 $\pm$ 0.002 & 0.032 $\pm$ 0.011 & 0.003 $\pm$ 0.002 \\
ObGD & 0.936 $\pm$ 0.030 & 0.644 $\pm$ 0.032 & 0.808 $\pm$ 0.069 & 0.306 $\pm$ 0.085 & 0.420 $\pm$ 0.191 & 0.113 $\pm$ 0.056 \\
AdaptiveObGD & \textbf{0.968} $\pm$ 0.011 & \textbf{0.747} $\pm$ 0.019 & 0.656 $\pm$ 0.092 & 0.181 $\pm$ 0.074 & 0.368 $\pm$ 0.044 & 0.097 $\pm$ 0.028 \\
AdaptiveObGD + CBP & 0.920 $\pm$ 0.025 & 0.697 $\pm$ 0.034 & 0.604 $\pm$ 0.059 & 0.246 $\pm$ 0.066 & 0.344 $\pm$ 0.173 & 0.147 $\pm$ 0.088 \\
AdaptiveObGD + LN & 0.864 $\pm$ 0.026 & 0.538 $\pm$ 0.011 & \textbf{0.848} $\pm$ 0.034 & \textbf{0.436} $\pm$ 0.046 & \textbf{0.676} $\pm$ 0.107 & \textbf{0.278} $\pm$ 0.026 \\
\hline
Batch PPO (No Warm-Start) & 0.492 $\pm$ 0.129 & 0.163 $\pm$ 0.044  & 0.088 $\pm$ 0.018 & 0.014 $\pm$ 0.005  & 0.040 $\pm$ 0.014 & 0.002 $\pm$ 0.001  \\
Batch PPO & 0.412 $\pm$ 0.154 & 0.111 $\pm$ 0.047  & 0.060 $\pm$ 0.020 & 0.008 $\pm$ 0.004 & 0.004 $\pm$ 0.009 & 0.000 $\pm$ 0.000 \\
\hline
Pretrained Policy & 0.060 & -  & 0.000 & - & 0.000 & - \\
Pretrained Policy (with LN) & 0.000 & - & 0.000 & - & 0.000 & -  \\
\hline
\end{tabular}
\vspace{-1.5em}
\label{table1}
\end{table*}

\subsection{Stream Adaptation Experiments}
\label{sec:analysis-main}
We analyze the impact of different optimizers and plasticity loss mitigation mechanisms with a total of 5 unique variants of Stream-AC. To compare the effects of using different optimizers, we have 3 variants that use Adam, ObGD, and AdaptiveObGD. The Adam variant uses a TD(0) Advantage Actor-Critic update, while ObGD and AdaptiveObGD use a TD($\lambda$) update. To compare plasticity loss mitigation mechanisms, we have 2 variants that both use the the AdaptiveObGD optimizer: one that includes continual backpropagation (CBP) and another that performs layer normalization (LN). Note that CBP is only incorporated during the stream learning phase. We used off-the-shelf default hyperparameters for ObGD and AdaptiveObGD variants, and we performed thorough hyperparameter tuning for the Adam variant until there was no severe performance collapse during the warm-start. 

These experiments model the non-stationary MDP $[M_0, M_1]$, where $M_0$ is the warm-start period on the base task and $M_1$ is the adaptation scenario lasting 1.5 million steps. During $M_1$, we obtain the max and mean success rates from the evaluations that occur periodically during training every 10,000 steps. The max success rate indicates how well the method is able to recover from the change, while the mean success rate shows the method's ability to maintain high performance throughout the adaptation scenario. Furthermore, to evaluate the stability of a policy during $M_1$, we use the mean absolute successive difference (MASD) metric formalized as $\frac{1}{N-1} \sum |x_n - x_{n-1}|$, where $N$ is the total number of sampled evaluations and $x_n$ is the success rate at the $n$th evaluation. Results can be found in Table \ref{table1}, Table \ref{table2}, and Fig. \ref{fig:goal_shift}.  

\subsection{Quantitative Results}
\label{sec:analysis-quant}
\begin{table}[ht]
    \centering
    \caption{Table 2. Average MASD $\pm$ STD Across 5 Seeds During Quadruped Locomotion Adaptation Phase \textit{(lower is more stable)}}
    \label{tab:masd_results}
    \begin{tabular}{l|c|c|c}
        \hline
        \textbf{Method} & \textbf{Broken Leg} & \textbf{Goal Shift} & \textbf{Slippery Floor} \\
        \hline
        Adam                    & 0.039 $\pm$ 0.00 & 0.024 $\pm$ 0.00 & 0.004 $\pm$ 0.00 \\
        ObGD                    & 0.124 $\pm$ 0.01 & 0.099 $\pm$ 0.02 & 0.059 $\pm$ 0.02 \\
        AdaptiveObGD            & 0.067 $\pm$ 0.00 & 0.046 $\pm$ 0.01 & 0.032 $\pm$ 0.01 \\
        \hline
        + CBP      & 0.067 $\pm$ 0.01 & 0.066 $\pm$ 0.01 & 0.045 $\pm$ 0.02 \\
        + LN       & 0.062 $\pm$ 0.00 & 0.063 $\pm$ 0.01 & 0.061 $\pm$ 0.01 \\
        \hline
    \end{tabular}
    \label{table2}
    \vspace{-0.5em}
\end{table}
\begin{figure}
    \includegraphics[width=\linewidth]{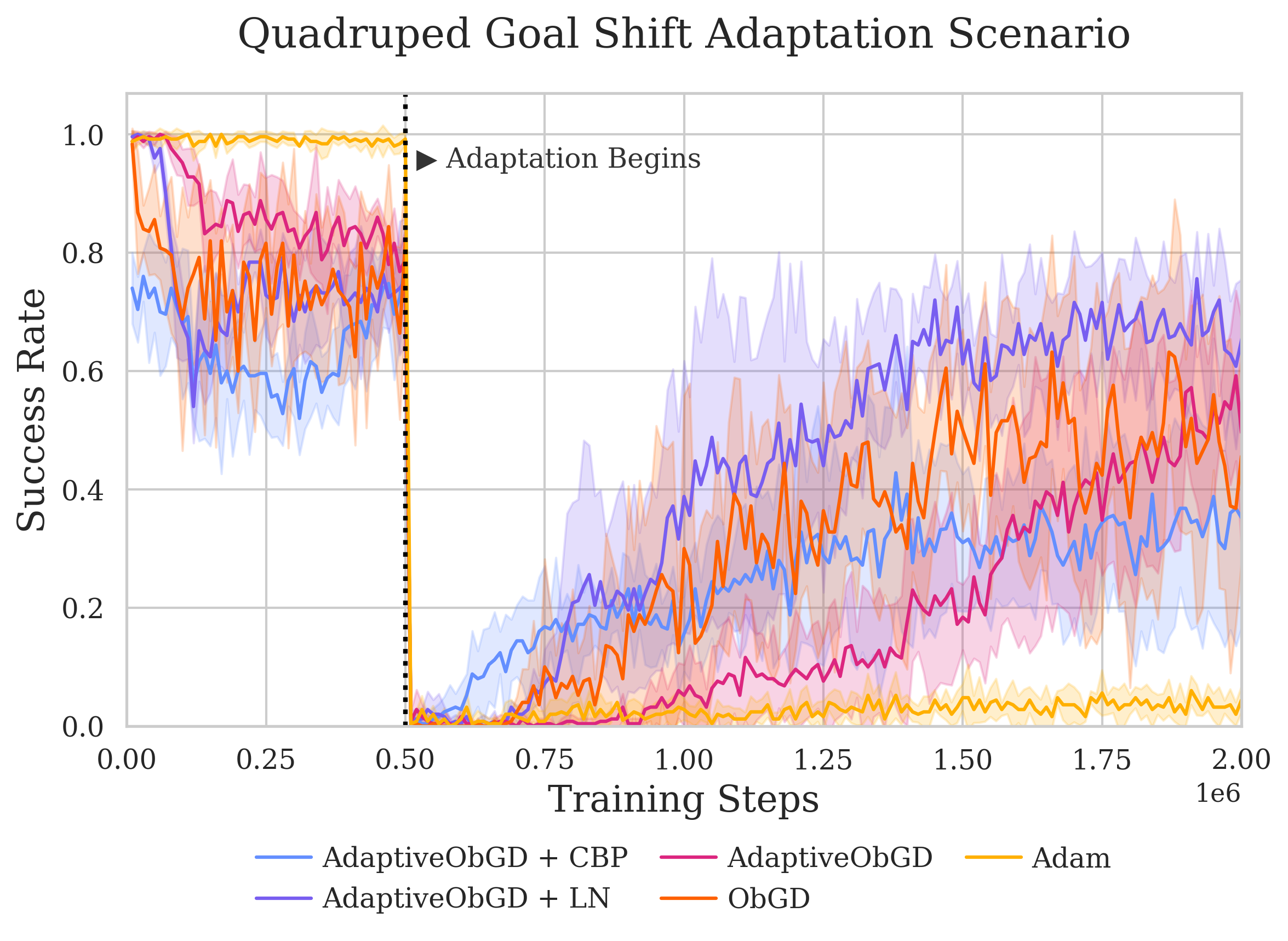}
    \caption{The performance of our different Stream-AC variants on the quadruped goal shift adaptation is shown. AdaptiveObGD + LayerNorm  outperforms other variants. ObGD is able to achieve a higher performance than AdaptiveObGD, but at the cost of stability.}
    \label{fig:goal_shift}
    \vspace{-1.8em}
\end{figure}


As shown in Table \ref{table1}, of the three optimizers tested, ObGD and AdaptiveObGD significantly outperformed Adam on all scenarios, featuring both a high max and mean succcess rate.  This result aligns with the literature suggesting that Adam is, in general, not well-suited for stream learning \cite{elsayed2024streamingdeepreinforcementlearning}, demonstrating that bounding mechanisms to prevent large step sizes and eligibility traces are critical for continual stream learning. Although, AdaptiveObGD slightly outperforms ObGD on the broken leg scenario by 3.2\% in max success rate, we find that it underperforms ObGD on goal shift by 15.2\% and slippery floor by 5.2\% on average. Taking a closer look at the MASD metric during adaptation (see Table \ref{table2}), we find that ObGD consistently has a higher MASD than AdaptiveObGD on average across all adaptation scenarios, indicating that ObGD is comparatively less stable than AdaptiveObGD. We can visually observe this result in Fig.~\ref{fig:goal_shift} where, despite the higher max and mean performance during the goal shift scenario, ObGD exhibits significantly greater performance instability during adaptation than AdaptiveObGD. This result suggests that having an adaptive optimizer with a second moment estimate can improve performance stability, although this may come at the cost of volatile performance spikes that can lead to a higher max and mean success rate.


With the plasticity loss mitigation techniques added to AdaptiveObGD, we see that including layer normalization results in the best performance on the goal shift and slippery floor adaptations of all Stream-AC variants. This conforms with prior work suggesting that LN is a beneficial addition for maintaining plasticity and improving training stability \cite{pmlr-v202-lyle23b,elsayed2024streamingdeepreinforcementlearning}. The results in Table \ref{table2} also show that, even with the notable performance boost of adding LN to AdaptiveObGD, it retains the performance stability benefits of the adaptive optimizer, producing a significantly lower MASD on the broken leg and goal shift scenario and only a slightly higher MASD on the slippery floor task.

Although the inclusion of CBP with AdaptiveObGD did not appear to significantly benefit the max success rate, we do find that AdaptiveObGD + CBP outperforms regular AdaptiveObGD in mean success rate on the goal shift and slippery floor scenarios. We interestingly observe in Fig.~\ref{fig:goal_shift} that, despite having the most significant performance degradation during the warm-start, AdaptiveObGD + CBP appears to be the first of all Stream-AC methods to meaningfully recover after the goal shift adaptation is introduced. The accelerated recovery likely explains why AdaptiveObGD + CBP has a higher mean success rate compared to regular AdaptiveObGD. This result highlights the potential of approaches that directly inject diversity into neural networks in accelerating adaptation, but also the long-standing challenge of managing the plasticity-stability tradeoff in deep learning.

\subsection{Comparison to Batch-Based Approaches}
To determine if there is a significant performance gap between streaming versus traditional batch deep RL, we evaluate our Stream-AC variants against a batch-based PPO implementation in our adaptation scenarios. We thoroughly tuned the hyperparameters of the batch PPO implementation (e.g., learning rate, batch size) to stabilize the policy during the warm-start, yet allow for meaningful recovery during the adaptation stage.

Surprisingly, our results shown in Table~\ref{table1} show that the top four Stream-AC variants, AdaptiveObGD, AdaptiveObGD + LN, AdaptiveObGD + CBP, and ObGD, are able to significantly outperform batch PPO on all adaptation scenarios in max and mean success rate. Because the literature suggests that Adam does not handle non-stationarity well \cite{pmlr-v202-lyle23b}, we compare our Stream-AC variants against a batch-based PPO that does not go through the warm-start period. Because of this change, the Adam optimizer does not have to handle any distribution shifts and we can set a slightly more aggressive learning rate to allow for larger parameter updates. Although this improves batch PPO's performance across all scenarios, thus corroborating the observed deficiencies with Adam in non-stationary settings, batch-based PPO still underperforms in success compared to our top performing Stream-AC variants on all tasks. 

These results underscore that, with careful design choices for the optimizer and plasticity-preserving components, streaming deep RL algorithms are adept at adapting pretrained policies to new scenarios, even when compared to common batch-based deep RL approaches.

\section{Analysis on Other Robotic Environments}
\label{sec:analysis2}
\begin{table*}[ht]
\centering
\caption{Table 3. Average Success Rates $\pm$ STD Across 5 Seeds on Manipulation Adaptation Scenarios}
\begin{tabular}{l|cc|cc}
\hline
\multirow{2}{*}{\textbf{Method}} & \multicolumn{2}{c|}{\textbf{Push Cube (Goal Shift)}} & \multicolumn{2}{c}{\textbf{Transport Box (Impaired Joints)}}  \\
 & Max & Mean & Max & Mean \\
\hline
ObGD & 0.396 $\pm$ 0.224 & 0.147 $\pm$ 0.084 & 0.132 $\pm$ 0.011& 0.048 $\pm$ 0.006\\
AdaptiveObGD & 0.496 $\pm$ 0.017 & \textbf{0.234} $\pm$ 0.008 & 0.148 $\pm$ 0.030& 0.059 $\pm$ 0.012\\
AdaptiveObGD + CBP & 0.396 $\pm$ 0.059 & 0.110 $\pm$ 0.017 & 0.152 $\pm$ 0.023& 0.067 $\pm$ 0.009\\
AdaptiveObGD + LN & \textbf{0.784} $\pm$ 0.046 & 0.233 $\pm$ 0.018 & \textbf{0.248} $\pm$ 0.030& \textbf{0.122} $\pm$ 0.014\\
\hline
Pretrained Policy & 0.028 & - & 0.080 & - \\
Pretrained Policy (with LN) & 0.000  & - & 0.080 & - \\
\hline
\end{tabular}
\vspace{-1.5em}
\label{table3}
\end{table*}
We conduct a series of experiments in robotic manipulation environments to see whether our previous observations about the performance of streaming deep RL for adaptation in quadruped locomotion hold with a different task space and robotic morphology. Note that we follow the same experimental setup as defined previously in \S\ref{sec:analysis-setup}.
\subsection{Task Environment}
Our subsequent analyses are performed on two object manipulation tasks with different robot embodiments in ManiSkill3 \cite{taomaniskill3}. For both environments, state space is used for observations and actions are represented by the change in joint positions. We also use the default dense reward function provided by the environment. 

\textbf{(1) Push Cube}: PushCube-v1 requires a Franka Emika Panda robot arm to push a cube to a goal position on a table. The cube's position on the table is randomized within a certain range in front of the robot, but the goal is always a set distance in front of the cube. The robot successfully completes the task if the cube is on the table and within a certain radius of the goal. The state space is comprised of joint position and velocity, the tool center point (TCP) pose, cube pose, and goal pose. The dense reward function is comprised of a distance to optimal push position reward, cube-to-goal distance reward, and a reward for keeping the cube on the table. 

\textbf{(2) Transport Box}: UnitreeG1TransportBox-v1 requires a Unitree G1 humanoid robot to move a box from one table to the other. The box's pose and orientation are randomized within a certain range on the table. The robot successfully completes the task if the box is resting on the other table.  The state space is comprised of joint position and velocity, TCP pose for the left and right arm, the box pose, and the distance between the right and left TCP pose to the box. The dense reward function consists of 4 stages encouraging the robot to: move to face the box, grasp the box stably, transport the box to the other table, and release the box.
\subsection{Adaptation Scenarios}
For each manipulation task, we developed a unique scenario that tests the robot's ability to adapt to abrupt changes to its embodiment or goals. \textbf{(1) Goal Shifts in Push Cube}: Although the cubes are still spawned within the same range, we consistently offset the goal position by -0.15 on the y-axis. \textbf{(2) Impaired Joints with Transport Box}: We make certain shoulder, hand, and finger joints stuck via zero-masking and stiff by scaling their values in the action vector.

\subsection{Stream Adaptation Experiments}
With this set of experiments, we evaluate if streaming deep RL can be used to adapt control policies to a single abrupt change to the robot or its goals within the context of an object manipulation task. We compare the top 4 performing Stream-AC variants from the quadruped locomotion experiments defined in \S\ref{sec:analysis-main}: ObGD, AdaptiveObGD, AdaptiveObGD + CBP, and AdaptiveObGD + LN. Although the actor-critic neural network architecture remains identical to that in the previous analysis, we were unable to directly apply the hyperparameters used previously, as all policies appeared prone to rapid divergence during the warm-start phase. Thus, this required us to perform extensive tuning of the scaling factor $\kappa$ for all of the evaluated methods. With the CBP variant, we also had to decrease the replacement rate $\rho$ to induce greater stability.



Similar to \S\ref{sec:analysis-main}, these experiments model the non-stationary MDP $[M_0, M_1]$, where $M_0$ is the warm-start period on the base task and $M_1$ is the adaptation scenario lasting 1.5 million steps. During $M_1$, we obtain the max and mean success rates from evaluations that occur every 10,000 steps. Results can be found in Table \ref{table3} and Fig.~\ref{fig:push_goal_shift}.  

\subsection{Quantitative Results}
\subsubsection{Push Cube}
\begin{figure}
    \includegraphics[width=\linewidth]{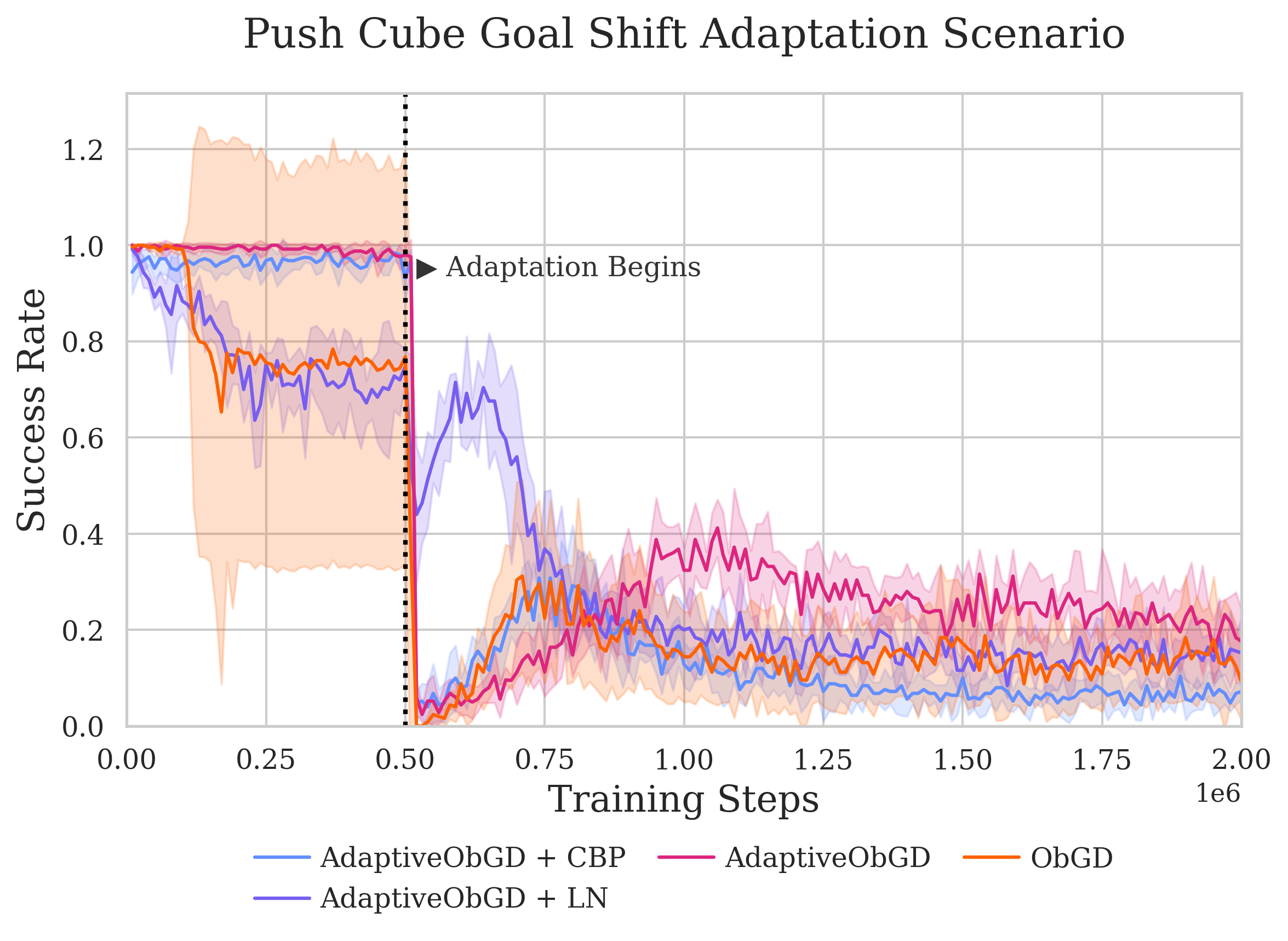}
    \caption{The performance of 4 Stream-AC variants on the Push Cube goal shift adaptation is shown. All methods show early signs of strong recovery, but performance declines over time after peaking.}
    \label{fig:push_goal_shift}
    \vspace{-1.8em}
\end{figure}
Our results show that streaming deep RL methods are able to adapt to goal shifts within the push cube task that would be catastrophic for the base policy. As shown in Table \ref{table3}, AdaptiveObGD + LN was able to achieve the highest max success rate at 78.4\% of the 4 approaches. This aligns with our previous results in quadruped locomotion, where the addition of layer normalization combined with AdaptiveObGD resulted in the best adaptation performance. Both the AdaptiveObGD + LN and AdaptiveObGD variants obtained the highest mean success rate at approximately 23\%. Taking a closer look at the success rate over time in Fig.~\ref{fig:push_goal_shift}, we observe a peculiar trend among all methods, where, after the adaptation scenario is introduced, they effectively recover to the method's respective max success rate before slowly declining to a stable lower success rate at around 20\%. Furthermore, the high variance in ObGD's warm-start performance can be attributed to its instability and occasional performance collapse even with extensive hyperparameter tuning. These results show a current limitation of streaming deep RL methods, where they appear capable of enabling neural networks to be sufficiently malleable to adapt an agent's behavior, but can still face significant performance decline and instability over time on manipulation tasks like Push Cube that may require greater precision.
\subsubsection{Transport Box}
Our results for the impaired joints adaptation for the transport box task, displayed in Table \ref{table3}, showed limited success across all variants during adaptation. The AdaptiveObGD + LN variant again demonstrated the strongest performance of all the compared methods with the highest max success rate at 23.3\% and mean success rate at 12.2\%.  AdaptiveObGD + CBP had the second strongest performance, narrowly outperforming the base AdaptiveObGD variant. Although the plasticity loss mitigation variants performed best, these performance improvements are marginal over the base pretrained policy. The humanoid robotic task setting is unique in its significantly larger action space, which is more than twice as large as the quadruped task. We believe that this result underscores another important limitation of current streaming deep RL methods where, on control tasks with greater complexity, there is still a limited capacity for performance recovery and behavioral adaptation from noisy single-sample stream updates. 
\section{Discussion}
\label{sec:discussion}
In this paper, we presented the first analysis of streaming deep reinforcement learning for adaptive continual learning in robotics. Our analysis was centered on using streaming deep RL to adapt pretrained policies online to abrupt changes to the robot's embodiment, environment, and goals. We show through our experiments that, with careful design considerations to optimizer choice and plasticity loss mitigation, streaming deep RL is effective in adapting pretrained policies to change, outperforming even common batch-based deep RL approaches like PPO.  

Our main takeaways are as follows:

\textbf{1. State-of-the-art streaming deep RL algorithms show sparks of effectiveness on certain robotic locomotion and manipulation tasks.}  We observe strong forward transfer capacity, especially within quadruped locomotion, in adapting policies to diverse scenarios. We witness signs of success in the manipulation domain but also the existing ceiling of stability and adaptability addressable by future work

\textbf{2. Design choices that manage the plasticity-stability tradeoff in neural networks are critical for performant adaptive stream learning.} These decisions can be made at the optimizer level, where carefully bounded updates, eligibility traces, and second moment estimates help induce stability. They also can be made at the level of the network's latent dynamics, where techniques like continual backpropagation and layer normalization preserve plasticity in both direct and indirect ways.

\textbf{3. Despite these promising results, realizing the full potential of stream learning in robotics remains a major open challenge.} While our experiments demonstrate encouraging progress in simulation, they represent only an early step toward the broader vision of truly in-situ robotic learning. In practice, robots deployed in the real world must operate over long horizons, continually adapting to evolving embodiments, environments, and objectives across a sequence of nonstationary MDPs. Achieving this level of persistent autonomy will require substantial advances in stability, safety, data efficiency, and system design. We view our work as an initial step toward this vision and hope it helps motivate further research toward robots capable of robust, lifelong adaptation during real-world deployment.

\bibliographystyle{plainnat}
\bibliography{refs}

@article{elsayed2024streamingdeepreinforcementlearning,
  title={Streaming Deep Reinforcement Learning Finally Works},
  author={Elsayed, Mohamed and Vasan, Gautham and Mahmood, A Rupam},
  journal={arXiv preprint arXiv:2410.14606},
  year={2024}
}

@inproceedings{
javed2024the,
title={The Big World Hypothesis and its Ramifications for Artificial Intelligence},
author={Khurram Javed and Richard S. Sutton},
booktitle={Finding the Frame: An RLC Workshop for Examining Conceptual Frameworks},
year={2024},
}

@article{dohare_loss_2024,
	title = {Loss of plasticity in deep continual learning},
	volume = {632},
	issn = {1476-4687},
	doi = {10.1038/s41586-024-07711-7},
	number = {8026},
	journal = {Nature},
	author = {Dohare, Shibhansh and Hernandez-Garcia, J. Fernando and Lan, Qingfeng and Rahman, Parash and Mahmood, A. Rupam and Sutton, Richard S.},
	month = aug,
	year = {2024},
	pages = {768--774},
}

@misc{intelligence2025pi05visionlanguageactionmodelopenworld,
      title={$\pi_{0.5}$: a Vision-Language-Action Model with Open-World Generalization}, 
      author={Physical Intelligence},
      year={2025},
      eprint={2504.16054},
      archivePrefix={arXiv},
      primaryClass={cs.LG},
}

@book{10.5555/3312046, author = {Sutton, Richard S. and Barto, Andrew G.}, title = {Reinforcement Learning: An Introduction}, year = {2018}, isbn = {0262039249}, publisher = {A Bradford Book}, address = {Cambridge, MA, USA} }

@inproceedings{andrychowicz_hindsight_2017,
	title = {Hindsight {Experience} {Replay}},
	volume = {30},
	urldate = {2025-08-31},
	booktitle = {Advances in {Neural} {Information} {Processing} {Systems}},
	author = {Andrychowicz, Marcin and Wolski, Filip and Ray, Alex and Schneider, Jonas and Fong, Rachel and Welinder, Peter and McGrew, Bob and Tobin, Josh and Pieter Abbeel, OpenAI and Zaremba, Wojciech},
	year = {2017},
}

@article{taomaniskill3,
  title={ManiSkill3: GPU Parallelized Robotics Simulation and Rendering for Generalizable Embodied AI},
  author={Stone Tao and Fanbo Xiang and Arth Shukla and Yuzhe Qin and Xander Hinrichsen and Xiaodi Yuan and Chen Bao and Xinsong Lin and Yulin Liu and Tse-kai Chan and Yuan Gao and Xuanlin Li and Tongzhou Mu and Nan Xiao and Arnav Gurha and Viswesh Nagaswamy Rajesh and Yong Woo Choi and Yen-Ru Chen and Zhiao Huang and Roberto Calandra and Rui Chen and Shan Luo and Hao Su},
  journal = {Robotics: Science and Systems},
  year={2025},
}

@inproceedings{vasan_deep_2024,
	title = {Deep {Policy} {Gradient} {Methods} {Without} {Batch} {Updates}, {Target} {Networks}, or {Replay} {Buffers}},
	volume = {37},
	booktitle = {Advances in {Neural} {Information} {Processing} {Systems}},
	author = {Vasan, Gautham and Elsayed, Mohamed and Azimi, Alireza and He, Jiamin and Shariar, Fahim and Bellinger, Colin and White, Martha and Mahmood, A. Rupam},
	year = {2024},
	pages = {845--891},
}

@article{mnih_human-level_2015,
	title = {Human-level control through deep reinforcement learning},
	volume = {518},
	issn = {1476-4687},
	doi = {10.1038/nature14236},
	number = {7540},
	journal = {Nature},
	author = {Mnih, Volodymyr and Kavukcuoglu, Koray and Silver, David and Rusu, Andrei A. and Veness, Joel and Bellemare, Marc G. and Graves, Alex and Riedmiller, Martin and Fidjeland, Andreas K. and Ostrovski, Georg and Petersen, Stig and Beattie, Charles and Sadik, Amir and Antonoglou, Ioannis and King, Helen and Kumaran, Dharshan and Wierstra, Daan and Legg, Shane and Hassabis, Demis},
	month = feb,
	year = {2015},
	pages = {529--533},
}

@misc{levine2020offlinereinforcementlearningtutorial,
      title={Offline Reinforcement Learning: Tutorial, Review, and Perspectives on Open Problems}, 
      author={Sergey Levine and Aviral Kumar and George Tucker and Justin Fu},
      year={2020},
      eprint={2005.01643},
      archivePrefix={arXiv},
      primaryClass={cs.LG},
}

@article{Elfwing2017SigmoidWeightedLU,
  title={Sigmoid-Weighted Linear Units for Neural Network Function Approximation in Reinforcement Learning},
  author={Stefan Elfwing and Eiji Uchibe and Kenji Doya},
  journal={Neural networks : the official journal of the International Neural Network Society},
  year={2017},
  volume={107},
  pages={
          3-11
        },
}

@InProceedings{pmlr-v202-lyle23b,
  title = 	 {Understanding Plasticity in Neural Networks},
  author =       {Lyle, Clare and Zheng, Zeyu and Nikishin, Evgenii and Avila Pires, Bernardo and Pascanu, Razvan and Dabney, Will},
  booktitle = 	 {Proceedings of the 40th International Conference on Machine Learning},
  pages = 	 {23190--23211},
  year = 	 {2023},
  volume = 	 {202},
  series = 	 {Proceedings of Machine Learning Research},
  month = 	 {23--29 Jul},
  publisher =    {PMLR}
}

@article{nikishin_deep_2023,
	title = {Deep {Reinforcement} {Learning} with {Plasticity} {Injection}},
	volume = {36},
	language = {en},
	urldate = {2025-09-05},
	journal = {Advances in Neural Information Processing Systems},
	author = {Nikishin, Evgenii and Oh, Junhyuk and Ostrovski, Georg and Lyle, Clare and Pascanu, Razvan and Dabney, Will and Barreto, Andre},
	month = dec,
	year = {2023},
	pages = {37142--37159},
}

@inproceedings{
muppidi2024fast,
title={Fast {TRAC}: A Parameter-Free Optimizer for Lifelong Reinforcement Learning},
author={Aneesh Muppidi and Zhiyu Zhang and Heng Yang},
booktitle={The Thirty-eighth Annual Conference on Neural Information Processing Systems},
year={2024},
}

@inproceedings{
ankile2024imitationrefinementresidual,
title={From Imitation to Refinement {\textendash} Residual {RL} for Precise Visual Assembly},
author={Lars Lien Ankile and Anthony Simeonov and Idan Shenfeld and Marcel Torne Villasevil and Pulkit Agrawal},
booktitle={CoRL Workshop on Learning Robot Fine and Dexterous Manipulation: Perception and Control},
year={2024},
}

@inproceedings{sun2025dynamic,
author = {Sun, Xiatao and Yang, Shuo and Chen, Yinxing and Fan, Francis and Liang, Yiyan and Rakita, Daniel},
year = {2025},
month = {06},
pages = {},
title = {Dynamic Rank Adjustment in Diffusion Policies for Efficient and Flexible Training},
doi = {10.15607/RSS.2025.XXI.159}
}

@inproceedings{DBLP:conf/rss/LyuLZZDW24,
  author={Shangke Lyu and Xin Lang and Han Zhao and Hongyin Zhang and Pengxiang Ding and Donglin Wang},
  title={RL2AC: Reinforcement Learning-based Rapid Online Adaptive Control for Legged Robot Robust Locomotion},
  year={2024},
  cdate={1704067200000},
  booktitle={Robotics: Science and Systems},
}

@article{dong2025batch,
    title   = {What Matters for Batch Online Reinforcement Learning in Robotics?},
    author  = {Perry Dong and Suvir Mirchandani and Dorsa Sadigh and Chelsea Finn},
    journal = {arXiv},
    year    = {2025},
}

@misc{schulman2017proximalpolicyoptimizationalgorithms,
      title={Proximal Policy Optimization Algorithms}, 
      author={John Schulman and Filip Wolski and Prafulla Dhariwal and Alec Radford and Oleg Klimov},
      year={2017},
      eprint={1707.06347},
      archivePrefix={arXiv},
      primaryClass={cs.LG},
}

@misc{ba2016layernormalization,
      title={Layer Normalization}, 
      author={Jimmy Lei Ba and Jamie Ryan Kiros and Geoffrey E. Hinton},
      year={2016},
      eprint={1607.06450},
      archivePrefix={arXiv},
      primaryClass={stat.ML},
}

@inproceedings{2015-kingma,
  author = {Kingma, Diederik P. and Ba, Jimmy},
  booktitle = {ICLR (Poster)},
  ee = {http://arxiv.org/abs/1412.6980},
  title = {Adam: A Method for Stochastic Optimization.},
  year = 2015
}

@ARTICLE{Hayes2021-ti,
  title     = "Replay in deep learning: Current approaches and missing
               biological elements",
  author    = "Hayes, Tyler L and Krishnan, Giri P and Bazhenov, Maxim and
               Siegelmann, Hava T and Sejnowski, Terrence J and Kanan,
               Christopher",
  journal   = "Neural Comput.",
  publisher = "MIT Press - Journals",
  volume    =  33,
  number    =  11,
  pages     = "2908--2950",
  month     =  oct,
  year      =  2021,
  language  = "en"
}

@article{lesort_continual_2020,
	title = {Continual learning for robotics: {Definition}, framework, learning strategies, opportunities and challenges},
	volume = {58},
	issn = {1566-2535},
	doi = {https://doi.org/10.1016/j.inffus.2019.12.004},
	journal = {Information Fusion},
	author = {Lesort, Timothée and Lomonaco, Vincenzo and Stoian, Andrei and Maltoni, Davide and Filliat, David and Díaz-Rodríguez, Natalia},
	year = {2020},
	pages = {52--68},
}

@inproceedings{octo_2023,
    title={Octo: An Open-Source Generalist Robot Policy},
    author = {{Octo Model Team}},
    booktitle = {Proceedings of Robotics: Science and Systems},
    address  = {Delft, Netherlands},
    year = {2024},
}


\end{document}